\documentclass{article}
\usepackage{ijcai20}

\usepackage{times}

\usepackage{soul}
\usepackage{url}
\usepackage[linkcolor=blue,breaklinks]{hyperref}
\usepackage[utf8]{inputenc}
\usepackage[small]{caption}
\usepackage{graphicx}
\usepackage{subfigure}
\usepackage{multirow}
\usepackage{amssymb}
\usepackage{amsmath}
\usepackage{booktabs}
\usepackage[ruled]{algorithm2e}
\usepackage{color}
\usepackage[switch]{lineno}

\title{Channel Pruning via Automatic Structure Search}

\author{
Mingbao Lin$^1$\and
Rongrong Ji$^1$\footnote{Corresponding Author}\and
Yuxin Zhang$^1$\and 
Baochang Zhang$^2$\and \\
Yongjian Wu$^3$\and
Yonghong Tian$^4$\\
\affiliations
$^1$Media Analytics and Computing Laboratory, Department of Artificial Intelligence, School of Informatics, Xiamen University, China\\
$^2$School of Automation Science and Electrical Engineering, Beihang University, China\\
$^3$Tencent Youtu Lab, Tencent Technology (Shanghai) Co., Ltd, China\\
$^4$School of Electronics Engineering and Computer Science, Peking University, Beijing, China\\
\emails
lmbxmu@stu.xmu.edu.cn,
rrji@xmu.edu.cn,
yxzhangxmu@163.com,
bczhang@buaa.edu.cn,
littlekenwu@tencent.com,
yhtian@pku.edu.cn
}

\begin{document}

\maketitle

\begin{abstract}
Channel pruning is among the predominant approaches to compress deep neural networks.
To this end, most existing pruning methods focus on selecting channels (filters) by importance/optimization or regularization based on rule-of-thumb designs, which defects in sub-optimal pruning.
In this paper, we propose a new channel pruning method based on artificial bee colony algorithm (ABC), dubbed as ABCPruner, which aims to efficiently find optimal pruned structure, \emph{i.e.}, channel number in each layer, rather than selecting ``important" channels as previous works did.
To solve the intractably huge combinations of pruned structure for deep networks,
we first propose to shrink the combinations where the preserved channels are limited to a specific space, thus the combinations of pruned structure can be significantly reduced.
And then, we formulate the search of optimal pruned structure as an optimization problem and integrate the ABC algorithm to solve it in an automatic manner to lessen human interference.
ABCPruner has been demonstrated to be more effective, which also enables the fine-tuning to be conducted efficiently in an end-to-end manner.
The source codes can be available at \url{https://github.com/lmbxmu/ABCPruner}.

\end{abstract}

\section{Introduction}\label{introduction}
The high demands in computing power and memory footprint of deep Convolutional Neural Networks (CNNs) prohibit their practical applications on edge computing devices such as smart phones or wearable gadgets.
To address this problem, extensive studies have made on compressing CNNs.
Prevalent techniques resort to quantization \cite{wang2019haq}, decomposition \cite{zhang2015efficient}, and pruning \cite{singh2019play}.
Among them, channel pruning has been recognized as one of the most effective tools for compressing CNNs \cite{luo2017thinet,he2017channel,he2018soft,liu2019metapruning,lin2020hrank}.

Channel pruning targets at removing the entire channel in each layer, which is straightforward but challenging because removing channels in one layer might drastically change the input of the next layer.
Most cutting-edge practice implements channel pruning by selecting channels (filters) based on rule-of-thumb designs.
Existing works follow two mainstreams.
The first pursues to identify the most important filter weights in a pre-trained model, which are then inherited by the pruned model as an initialization for the follow-up fine-tuning \cite{hu2016network,li2017pruning,he2017channel,he2018soft}.
It usually performs layer-wise pruning and fine-tuning, or layer-wise weight reconstruction followed by a data-driven and/or iterative optimization to recover model accuracy, both of which however are time-cost.
The second typically performs channel pruning based on handcrafted rules to regularize the retraining of a full model followed by pruning and fine-tuning \cite{liu2017learning,huang2018data,lin2019towards}.
It requires human experts to design and decide hyper-parameters like sparsity factor and pruning rate,
which is not automatic and thus less practical in compressing various CNN models.
Besides, rule-of-thumb designs usually produce the sub-optimal pruning \cite{he2018amc}.

The motivation of our ABCPruner is two-fold.
First, \cite{liu2019rethinking} showed that the essence of channel pruning lies in finding optimal pruned structure, \emph{i.e.}, channel number in each layer, instead of selecting ``important" channels.
Second, \cite{he2018amc} proved the feasibility of applying automatic methods for controlling hyper-parameters to channel pruning, which requires less human interference.

However, exhaustively searching for the optimal pruning structure is of a great challenge.
Given a CNN with $L$ layers, the combinations of pruned structure could be $\prod_{j=1}^Lc_j$, where $c_j$ is channel number in the $j$-th layer.
The combination overhead is extremely intensive for deep CNNs\footnote{For example, the combinations for VGGNet-16 are $\text{2}^{\text{104}}$, $\text{2}^{\text{448}}$ for GoogLeNet, and $\text{2}^{\text{1182}}$ for ResNet-152.}, which is prohibitive for resource-limited scenarios.
%

\begin{figure}[!t]
\begin{center}
\includegraphics[height=0.43\linewidth]{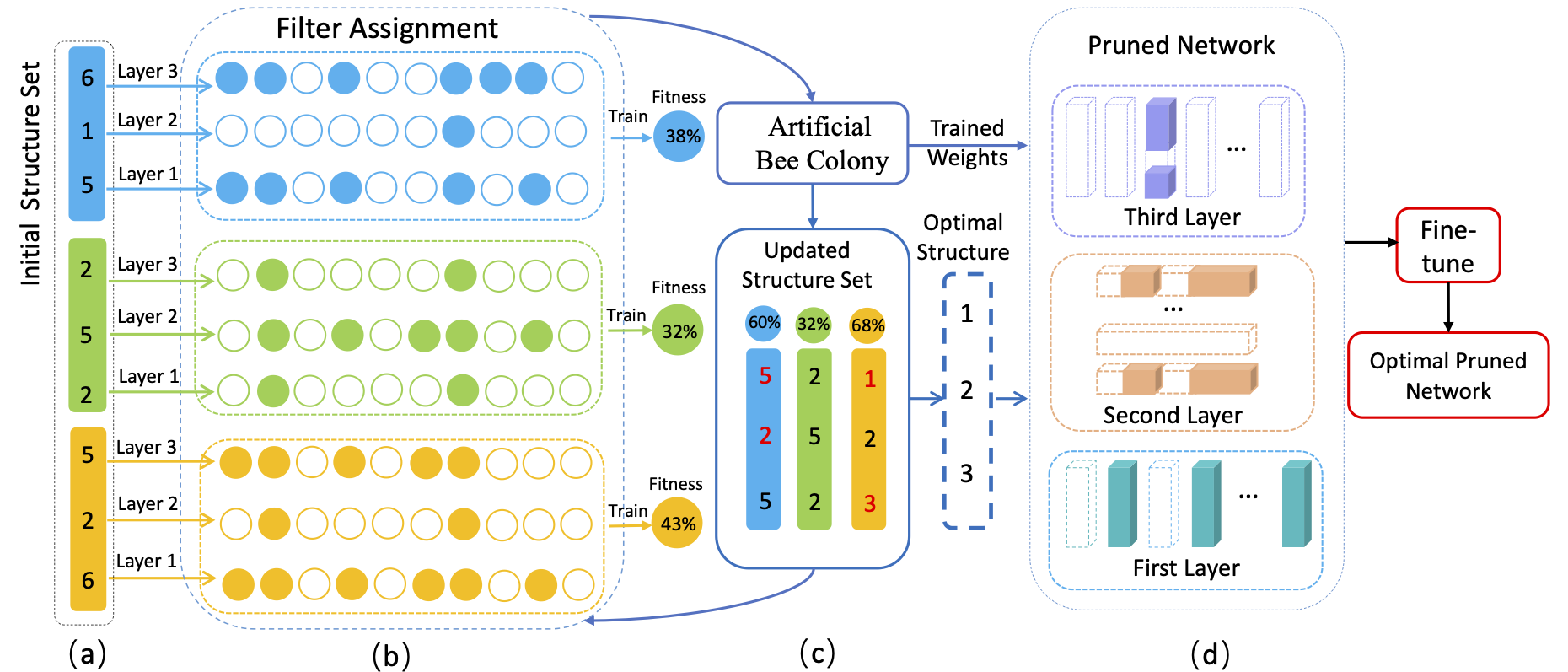}
\end{center}
\caption{\label{framework}
Framework of ABCPruner.
(a) A structure set is initialized first, elements of which represent the preserved channel number.
(b) The filters of the pre-trained model are randomly assigned to each structure.
We train it for given epochs to measure its fitness.
(c) Then, the ABC algorithm is introduced to update the structure set and the fitness is recalculated through (b).
(b) and (c) will continue for some cycles.
(d) The optimal pruned structure with best fitness is picked up, and the trained weights are reserved as a warm-up for fine-tuning the pruned network. (Best viewed with zooming in)
}
\end{figure}

%
In this paper, we introduce ABCPruner towards optimal channel pruning. 
To solve the above problem, two strategies are adopted in our ABCPruner.
First, we propose to shrink the combinations by limiting the number of preserved channels to $\{0.1c_j, 0.2c_j, ..., \alpha c_j\}$ where the value of $\alpha$ falls in $\{10\%, 20\%, ..., 100\%\}$, which shows that there are $10\alpha$ feasible solutions for each layer, and up to $\alpha \%$ percentage of channels in each layer will be preserved.
This operation is interpretable since channel pruning usually preserves a certain percentage of channels.
By doing so, the combinations will be significantly reduced to $(10{\alpha})^L \ll \prod_{j=1}^Lc_j$, making the search more efficient.
Second, we obtain the optimal pruned structure through the artificial bee colony (ABC) algorithm \cite{karaboga2005idea}.
ABC is automatic-based rather than hand-crafted, which thus lessens the human interference in determining the channel number.
As shown in Fig.\,\ref{framework}, we first initialize a structure set, each element of which represents the preserved channel number in each layer.
The filter weights of the full model are randomly selected and assigned to initialize each structure.
We train it for a given number of epochs to measure its fitness, \emph{a.k.a}, accuracy performance in this paper.
Then, ABC is introduced to update the structure set.
Similarly, filter assignment, training and fitness calculation for the updated structures are conducted.
We continue the search for some cycles.
Finally, the one with the best fitness is considered as the optimal pruned structure, and its trained weights are reserved as a warm-up for fine-tuning.
Thus, our ABCPruner can effectively implement channel pruning via automatically searching for the optimal pruned structure.

%
%
To our best knowledge, there exists only one prior work \cite{liu2019metapruning} which considers pruned structure in an automatic manner.
Our ABCPruner differs from it in two aspects:
First, the pruning in \cite{liu2019metapruning} is a two-stage scheme where a large PruningNet is trained in advance to predict weights for the pruned model, and then the evolutionary algorithm is applied to generate a new pruned structure.
While our ABCPruner is one-stage without particularly designed network, which thus simplifies the pruning process.
Second, the combinations in ABCPruner are drastically reduced to $(10{\alpha})^L$ ($\alpha \in \{10\%, 20\%, ..., 100\%\}$), which provides more efficiency for the channel pruning.

\section{Related Work}\label{related}
%
%
%

\paragraph{Network Pruning.}
Network pruning can be categorized into either weight pruning or channel pruning.
Weight pruning removes individual neurons in the filters or connections across different layers \cite{han2015learning,guo2016dynamic,li2017pruning,aghasi2017net,zhu2017prune,frankle2019lottery}.
After pruning, a significant portion of CNNs' weights are zero, and thus the memory cost can be reduced by arranging the model in a sparse format.
However, such a pruning strategy usually leads to an irregular network structure and memory access, which requires customized hardware and software to support practical speedup.

In contrast, channel pruning is especially advantageous by removing the entire redundant filters directly, which can be well supported by general-purpose hardware and BLAS libraries.
Existing methods pursue pruning by channel selection based on rule-of-thumb designs.
To that effect,
many works aim to inherit channels based on the importance estimation of filter weights, \emph{e.g.}, $l_p$-norm \cite{li2017pruning,he2018soft} and sparsity of activation outputs \cite{hu2016network}.
\cite{luo2017thinet,he2017channel} formulated channel pruning as an optimization problem, which selects the most representative filters to recover the accuracy of pruned network with minimal reconstruction error.
However, it is still hard to implement in an end-to-end manner without iteratively pruning (optimization) and fine-tuning.
Another group focuses on regularized-based pruning.
For example, \cite{liu2017learning} imposed sparsity constraint on the scaling factor of batch normalization layer,
while \cite{huang2018data,lin2019towards} proposed a sparsity-regularized mask for channel pruning, which is optimized through a data-driven selection or generative adversarial learning.
However, these methods usually require another round of retraining and manual hyper-parameter analysis.
Our work differs from traditional methods in that it is automatic and the corresponding fine-tuning is end-to-end, which has been demonstrated to be feasible by the recent work in \cite{he2018amc}.

\paragraph{AutoML.}
Traditional pruning methods involve human expert in hyper-parameter analysis, which hinders their practical applications.
Thus, automatic pruning has attracted increasing attention, which can be regarded as a specific AutoML task.
Most prior AutoML based pruning methods are implemented in a bottom-up and layer-by-layer manner, which typically achieve the automations through Q-value \cite{lin2017runtime}, reinforcement learning \cite{he2018amc} or an automatic feedback loop \cite{yang2018netadapt}.
Other lines include, but not limited to, constraint-aware optimization via an annealing strategy \cite{chen2018constraint} and sparsity-constraint regularization via a joint training manner \cite{luo2018autopruner}.
Different from most prior AutoML pruning, our work is inspired by the recent work in  \cite{liu2019rethinking} which reveals that the key of channel pruning lies in the pruned structure, \emph{i.e.}, channel number in each layer, instead of selecting ``important" channels.

\section{The Proposed ABCPruner}\label{abcpruner}

Given a CNN model $\mathcal{N}$ that contains $L$ convolutional layers and its filters set $\mathbf{W}$,
we refer to $C = ( c_1, c_2, ..., c_L )$  as the network structure of $\mathcal{N}$, where $c_j$ is the channel number of the $j$-th layer.
Channel pruning aims to remove a portion of filters in $\mathbf{W}$ while keeping a comparable or even better accuracy.
To that effect, traditional methods focus on selecting channels (filters) based on rule-of-thumb designs, which are usually sub-optimal \cite{he2018amc}.
Instead, our ABCPruner differs from these methods by finding the optimal pruned structure, \emph{i.e.}, the channel number in each layer,
and then solves the pruning in an automatic manner by integrating off-the-shelf ABC algorithm \cite{karaboga2005idea}.
We illustrate our ABCPruner in Fig.\,\ref{framework}.

\subsection{Optimal Pruned Structure}\label{optimal}

Our channel pruning is inspired by the recent study in \cite{liu2019rethinking} which reveals that the key step in channel pruning lies in finding the optimal pruned structure, \emph{i.e.}, channel number in each layer, rather than selecting ``important" channels.

For any pruned model $\mathcal{N}'$, we denote its structure as $C' = ( c'_1, c'_2, ..., c'_L )$, where $c'_j \le c_j$ is the channel number of the pruned model in the $j$-th layer.
Given the training set $\mathcal{T}_{train}$ and test set $\mathcal{T}_{test}$,
we aim to find the optimal combination of $C'$, such that the pruned model $\mathcal{N}'$ trained/fine-tuned on $\mathcal{T}_{train}$ obtains the best accuracy.
To that effect, we formulate our channel pruning problem as
\begin{equation}\label{optimization}
(C')^* = \mathop{\arg\max}\limits_{C'} \; acc\big(\mathcal{N}'(C', \mathbf{W}';\mathcal{T}_{train}); \mathcal{T}_{test}\big),
\end{equation}
where $\mathbf{W}'$ is the weights of pruned model trained/fine-tuned on $\mathcal{T}_{train}$, and $acc(\cdot)$ denotes the accuracy on $\mathcal{T}_{test}$ for $\mathcal{N}'$ with structure $C'$.
As seen, our implementation is one-stage where the pruned weights are updated directly on the pruned model, which differs from \cite{liu2019metapruning} where an extra large PruningNet has to be trained to predict weights for $\mathcal{N}'$, resulting in high complexity in the channel pruning.
However, the optimization of Eq.\,\ref{optimization} is almost intractable.
Detailedly, the potential combinations of network structure $C'$ are extremely large.
Exhaustively searching for Eq.\,\ref{optimization} is infeasible.
To solve this problem, we further propose to shrink the combinations in Sec.\,\ref{search}.

\subsection{Combination Shrinkage}\label{search}

%
%
Given a network $\mathcal{N}$, the combination of pruned structure could be $\prod_{i=1}^Lc_i$, which are extremely large and computationally prohibitive for the resource-limited devices.
Hence, we further propose to constrain Eq.\,\ref{optimization} as:
\begin{equation}\label{shrinkage}
\begin{split}
(C')^*  = & \mathop{\arg\max}\limits_{C'} \;  acc\big(\mathcal{N}'(C', \mathbf{W}';\mathcal{T}_{train}); \mathcal{T}_{test}\big),
\\&
s.t. \;\; c'_i \in \{ 0.1c_i, 0.2c_i, ..., \alpha c_i \}^L,
\end{split}
\end{equation}
where $\alpha = 10\%, 20\%, ..., 100\%$ is a pre-given constant and its influence is analyzed in Sec.\,\ref{alpha_influence}.
It denotes that for the $i$-th layer, at most $\alpha$ percentage of the channels in the pre-trained network $\mathcal{N}$ are preserved in $\mathcal{N}'$, \emph{i.e.}, $c'_i \le \alpha c_i $, and the value of $c'_i$ is limited in $\{0.1c_i, 0.2c_i, ..., \alpha_ic_i \}$ \footnote{When $ \alpha c_i  < 1$, we set $c'_i = 1$ to preserve one channel.}.
Note that $\alpha$ is shared across all layers, which greatly relieves the burden of human interference to involve in the parameter analysis.

Our motivations of introducing $\alpha$ are two-fold:
First, typically, a certain percentage of filters in each layer are preserved in the channel pruning, and $\alpha$ can play as the upbound of preserved filters.
Second, the introduced $\alpha$ significantly decreases the combinations to $(10{\alpha})^L$ ($\alpha \le$ 1), which makes solving Eq.\,\ref{optimization} more feasible and efficient.

Though the introduction of $\alpha$ significantly reduces the combinations of the pruned structure, it still suffers large computational cost to enumerate the remaining $(10\alpha)^L$ combinations for deep neural networks.
One possible solution is to rely on empirical settings, which however requires manual interference.
To solve the problem, we further automate the structure search in Sec.\,\ref{automatic}.

\begin{algorithm}[!t]
\caption{\label{alg1}ABCPruner}

\LinesNumbered
\KwIn{Cycles: $\mathcal{T}$, Upbound: $\alpha$, Number of pruned structures: $n$, Counter: $\{t_i\}_{i=1}^n$ = 0, Max times: $\mathcal{M}$.}
\KwOut{Optimal pruned structure $(C')^*$.}
Initialize the pruned structure set $\{C'_j\}_{j=1}^n$\;
\For{$h = 1 \rightarrow \mathcal{T}$}
{

\For{$j = 1 \rightarrow n$}
{
    generate candidate $G'_j$ via Eq.\,\ref{food_update}\;
    calculate the fitness of $C'_j$ and $G'_j$ via Eq.\,\ref{fit}\;
    \uIf{${fit}_{C'_j} < {fit}_{G'_j}$}
    {
       $C'_j = G'_j$\;
       ${fit}_{C'_j} = {fit}_{G'_j}$\;
       $t_j = 0$\;
    }\Else{$t_j$ = $t_j$ + 1\;}
}

\For{$j = 1 \rightarrow n$}
{
    calculate the probability of $P_j$ via Eq.\,\ref{probability}\;
    generate a random number ${\epsilon}_j \in [0, 1]$\;
    \If{${\epsilon}_j <= P_j$}
    {
        generate candidate $G'_j$ via Eq.\,\ref{food_update}\;
        calculate the fitness of $G'_j$ via Eq.\,\ref{fit}\;
        \uIf{${fit}_{C'_j} < {fit}_{G'_j}$}
       {
          $C'_j = G'_j$\;
          ${fit}_{C'_j} = {fit}_{G'_j}$\;
          $t_j = 0$\;
       }\Else{$t_j$ = $t_j$ + 1\;}

    }
}

\For{$j = 1 \rightarrow n$}
{
   \If{$t_j > \mathcal{M}$ }
   {
      re-initialize $C'_j$\;
   }
}

}
$(C')^* = \mathop{\arg\max}\limits_{C'_j} \;  acc\big(\mathcal{N}'_j(C'_j, \mathbf{W}'_j;\mathcal{T}_{train}); \mathcal{L}_{test}\big)$.
\end{algorithm}

\subsection{Automatic Structure Search}\label{automatic}
Instead of resorting to manual structure design or enumeration of all potential structure (channel number in each layer),
we propose to automatically search the optimal structure.

In particular, we initialize a set of $n$ pruned structures $\{ C'_j \}_{j=1}^n$ with the $i$-th element $c'_{ji}$ of $C'_j$ randomly sampled from $\{0.1c_i, 0.2c_i, ..., \alpha c_i\}$.
Accordingly, we obtain a set of pruned model $\{ \mathcal{N}'_j \}_{j=1}^n$ and a set of pruned weights $\{ \mathbf{W}'_j \}_{j=1}^n$.
Each pruned structure $C'_j$ represents a potential solution to the optimization problem of Eq.\,\ref{shrinkage}.
Our goal is to progressively modify the structure set and finally pick up the best pruned structure.
To that effect, we show that our optimal structure search for channel pruning can be optimized automatically by integrating off-the-shelf ABC algorithm \cite{karaboga2005idea}.
We first outline our algorithm in Alg.\,\ref{alg1}. More details are elaborated below, which mainly contains three steps.
%

%

%
\paragraph{Employed Bee (Line 3 -- Line 13).} 
The employed bee generates a new structure candidate $G'_j$ for each pruned structure $C'_j$.
The $i$-th element of $G'_j$ is defined as below:
\begin{equation}\label{food_update}
g'_{ji} = \lceil c'_{ji} + r \cdot ( c'_{ji} - c'_{gi} ) \rfloor,
\end{equation}
where $r$ is a random number within $[-1, +1]$ and $g \neq j$ denotes the $g$-th pruned structure.
$\lceil \cdot \rfloor$ returns the value closest to the input in $\{0.1c_i, 0.2c_i, ..., \alpha c_i\}$.

Then, employed bee will decide if the generated candidate $G'_j$ should replace $C'_j$ according to their fitness, which is defined in our implementation as below:
\begin{equation}\label{fit}
fit_{C'_j} = acc\big(\mathcal{N}'_j(C'_j, \mathbf{W}'_j;\mathcal{T}_{train}); \mathcal{T}_{test}\big).
\end{equation}

If the fitness of $G'_j$ is better than $C'_j$, $C'_j$ will be updated as $C'_j = G'_j$.
Otherwise, $C'_j$ will keep unchanged.

\paragraph{Onlooker Bee (Line 14 -- Line 28).} 
The onlooker bee further updates $\{ C_j' \}_{j=1}^n$ following the rule of Eq.\,\ref{food_update}.
Differently, a pruned structure $C'_j$ is chosen with a probability related to its fitness, defined as:
\begin{equation}\label{probability}
P_j = 0.9 \cdot \frac{fit_{C'_j}}{\max(fit_{C'_j})} + 0.1.
\end{equation}

Therefore, the better fitness of $C'_j$ is, the higher probability of $C'_j$ would be selected, which then produces a new and better pruned structure.
By this, ABCPruner will automatically result in the optimal pruned structure progressively.
\paragraph{Scout Bee (Line 29 -- Line 33).}
If a pruned structure $C'_j$ has not been updated more than $\mathcal{M}$ times, the scout bee will re-initialize it to further produce a new pruned structure.

However, to calculate the fitness in Eq.\,\ref{fit}, it has to train $\mathcal{N}_j'$ on $\mathcal{T}_{train}$ which is time-consuming and infeasible, especially when $\mathcal{T}_{train}$ is large-scale.
To solve this problem, as shown in Fig.\,\ref{framework}, given a potential combination of $C'$, we first randomly pick up $c'_j$ filters from the pre-trained model, which then serve as the initialization for the $j$-th layer of the pruned model $\mathcal{N}'$.
Then, we train $\mathcal{N}'$ for some small epochs to obtain its fitness.
Lastly, we give more epochs to fine-tune the pruned model with the best structure $(C')^*$, \emph{a.k.a.}, best fitness.
More details about the fine-tuning are discussed in Sec.\,\ref{implementation}.

%
%

\section{Experiments}\label{experiments}
We conduct compression for representative networks, including VGGNet, GoogLeNet and ResNet-56/110 on CIFAR-10 \cite{krizhevsky2009learning}, and ResNet-18/34/50/101/152 on ILSVRC-2012 \cite{russakovsky2015imagenet}.

\begin{table*}[!t]
\centering
\setlength{\tabcolsep}{0.48em}
\begin{tabular}{ccccccccc}
\toprule
Model    &Top1-acc &$\uparrow\downarrow$ &Channel  &Pruned &FLOPs  &Pruned  &Parameters &Pruned\\ \hline
VGGNet-16 Base     &93.02\%    &0.00\%            &4224  &0.00\%    &314.59M  &0.00\% &14.73M     &0.00\%    \\
\textbf{VGGNet-16 ABCPruner-80\%}   &93.08\%    &0.06\%$\uparrow$  &1639  &61.20\% &82.81M   &73.68\%     &1.67M  &88.68\% \\
\midrule
GoogLeNet Base &95.05\%  &0.00\%               &7904  &0.00\% &1534.55M &0.00\% &6.17M   &0.00\%    \\
\textbf{GoogLeNet ABCPruner-30\%} &94.84\%    &0.21\%$\downarrow$ &6150  &22.19\% &513.19M &66.56\%  &2.46M  &60.14\% \\
\midrule
ResNet-56 Base  &93.26\%    &0.00\%   &2032 &0.00\%  &127.62M  &0.00\% &0.85M    &0.00\%    \\
\textbf{ResNet-56 ABCPruner-70\%} &93.23\%    &0.03\%$\downarrow$  &1482 &27.07\% &58.54M&54.13\%  &0.39M   &54.20\%\\
\midrule
ResNet-110 Base &93.50\%    &0.00\%     &4048   &0.00\% &257.09M  &0.00\% &1.73M   &0.00\%    \\
\textbf{ResNet-110 ABCPruner-60\%} &93.58\%    &0.08\%$\uparrow$  &2701  &33.28\%  &89.87M &65.04\% &0.56M  &67.41\%\\
\bottomrule
\end{tabular}
\caption{\label{cifar}Accuracy and pruning ratio on CIFAR-10. We count the pruned channels, parameters and FLOPs for VGGNet-16 \protect\cite{simonyan2015very}, GoogLeNet \protect\cite{szegedy2015going} and ResNets with different depths of 56 and 110 \protect\cite{he2016deep}. ABCPruner-$\alpha$ here denotes that at most $\alpha$ percentage of channels are preserved in each layer.}
\end{table*}

\begin{table*}[!t]
\centering
\setlength{\tabcolsep}{0.125em}
\begin{tabular}{ccccccccccc}
\toprule
Model   &Top1-acc &$\uparrow\downarrow$ &Top5-acc &$\uparrow\downarrow$ &Channel &Pruned  &FLOPs  &Pruned  &Parameters &Pruned \\ \hline
ResNet-18 Base          &69.66\%&0.00\%&89.08\%&0.00\%&4800&0.00\% &1824.52M&0.00\% &11.69M&0.00\% \\
\textbf{ResNet-18 ABCPruner-70\%}&67.28\%&2.38\%$\downarrow$&87.67\%&1.41\%$\downarrow$&3894&18.88\%&1005.71M&44.88\%&6.60M&43.55\% \\
\textbf{ResNet-18 ABCPruner-100\%}&67.80\%&1.86\%$\downarrow$&88.00\%&1.08\%$\downarrow$&4220&12.08\%&968.13M&46.94\%&9.50M&18.72\% \\
\midrule
ResNet-34 Base           &73.28\%&0.00\%&91.45\%&0.00\%&8512&0.00\%&3679.23M&0.00\%&21.90M&0.00\% \\
\textbf{ResNet-34 ABCPruner-50\%} &70.45\%&2.83\%$\downarrow$&89.69\%&1.76\%$\downarrow$&6376&25.09\%&1509.76M&58.97\%&10.52M&51.76\% \\
\textbf{ResNet-34 ABCPruner-90\%} &70.98\%&2.30\%$\downarrow$&90.05\%&1.40\%$\downarrow$&6655&21.82\%&2170.77M&41.00\%&10.12M&53.58\% \\
\midrule
ResNet-50 Base           &76.01\%&0.00\%&92.96\%&0.00\%&26560&0.00\%&4135.70M&0.00\%&25.56M&0.00\% \\
\textbf{ResNet-50 ABCPruner-70\%} &73.52\%&2.49\%$\downarrow$&91.51\%&1.45\%$\downarrow$&22348&15.86\%&1794.45M&56.61\%&11.24M&56.01\% \\
\textbf{ResNet-50 ABCPruner-80\%} &73.86\%&2.15\%$\downarrow$&91.69\%&1.27\%$\downarrow$&22518&15.22\%&1890.60M&54.29\%&11.75M&54.02\% \\
\midrule
ResNet-101 Base          &77.38\%&0.00\%&93.59\%&0.00\%&52672&0.00\%&7868.40M&0.00\%&44.55M&0.00\% \\
\textbf{ResNet-101 ABCPruner-50\%}&74.76\%&2.62\%$\downarrow$&92.08\%&1.51\%$\downarrow$&41316&21.56\%&1975.61M&74.89\%&12.94M&70.94\% \\
\textbf{ResNet-101 ABCPruner-80\%}&75.82\%&1.56\%$\downarrow$&92.74\%&0.85\%$\downarrow$&43168&17.19\%&3164.91M&59.78\%&17.72M&60.21\% \\
\midrule
ResNet-152 Base          &78.31\%&0.00\%&93.99\%&0.00\%&75712&0.00\%&11605.91M&0.00\%&60.19M&0.00\% \\
\textbf{ResNet-152 ABCPruner-50\%}&76.00\%&2.31\%$\downarrow$&92.90\%&1.09\%$\downarrow$&58750&22.40\%&2719.47M&76.57\%&15.62M&74.06\%\\
\textbf{ResNet-152 ABCPruner-70\%}&77.12\%&1.19\%$\downarrow$&93.48\%&0.51\%$\downarrow$&62368&17.62\%&4309.52M&62.87\%&24.07M&60.01\%\\
\bottomrule
\end{tabular}
\caption{\label{imagenet}Accuracy and pruning ratio on ILSVRC-2012. We count pruned channels, parameters and FLOPs for ResNets with different depths of 18, 34, 50, 101 and 152 \protect\cite{he2016deep}. ABCPruner-$\alpha$ here denotes that at most $\alpha$ percentage of channels are preserved in each layer.}
\end{table*}

\subsection{Implementation Details}\label{implementation}
\paragraph{Training Strategy.}
We use the Stochastic Gradient Descent algorithm (SGD) for fine-tuning with momentum 0.9 and the batch size is set to 256.
On CIFAR-10, the weight decay is set to 5e-3 and we fine-tune the network for 150 epochs with a learning rate of 0.01, which is then divided by 10 every 50 training epochs.
On ILSVRC-2012, the weight decay is set to 1e-4 and 90 epochs are given for fine-tuning.
The learning rate is set as 0.1, and divided by 10 every 30 epochs.
%

%
\paragraph{Performance Metric.}
Channel number, FLOPs (floating-point operations), and parameters are used to measure the network compression.
%
%
Besides, for CIFAR-10, top-1 accuracy of pruned models are provided.
For ILSVRC-2012, both top-1 and top-5 accuracies are reported.

For each structure, we train the pruned model $\mathcal{N}'$ for two epochs to obtain its fitness.
We empirically set $\mathcal{T}$=2, $n$=3, and $\mathcal{M}$=2 in the Alg.\,\ref{alg1}.

\subsection{Results on CIFAR-10}\label{results_cifar}
We conduct our experiments on CIFAR-10 with three classic deep networks including VGGNet, GoogLeNet and ResNets.
The experiment results are reported in Tab.\,\ref{cifar}.

\paragraph{VGGNet.}
For VGGNet, the 16-layer (13-Conv + 3FC) model is adopted for compression on CIFAR-10.
We remove 61.20\% channels, 73.68\% FLOPs and 88.68\% parameters while still keeping the accuracy at 93.08\%, which is even slightly better than the baseline model.
This greatly facilitates VGGNet model, a popular backbone for object detection and semantic segmentation, to be deployed on mobile devices.

\paragraph{GoogLeNet.}
For GoogLeNet, as can be seen from Tab.\,\ref{cifar}, we remove 22.19\% inefficient channels with only 0.21\% performance drop of accuracy.
Besides, nearly 66.56\% FLOPs computation is saved and 60.14\% parameters are reduced.
Therefore, ABCPruner can be well applied even on the compact-designed GoogLeNet.
%
%
%

%
\paragraph{ResNets.}
For ResNets, we choose two different depths, including ResNet-56 and ResNet-110.
%
%
As seen from Tab.\,\ref{cifar}, the pruning rates with regard to channels, FLOPs and parameters increase from ResNet-56 to ResNet-110.
Detailedly, the channels reduction raises from 27.07\% to 33.28\%;
the FLOPs reduction raises from 54.13\% to 65.05\%;
and the parameters reduction raises from 54.20\% to 67.41\%.
To explain, ResNet-110 is deeper and more over-parameterized.
ABCPruner can automatically find out the redundancies and remove them.
Besides, it retains comparable accuracies against the baseline model, which verifies the efficacy of ABCPruner in compressing the residual-designed networks.

%

\subsection{Results on ILSVRC-2012}\label{results_ilsvrc}
We further perform our method on the large-scale ILSVRC-2012 for ResNets with different depths, including 18/34/50/101/152.
We present two different pruning rates for each network and the results are reported in Tab.\,\ref{imagenet}.

From Tab.\,\ref{imagenet}, we have two observations.
First observation is that the performance drops on ILSVRC-2012 are more than these on CIFAR-10.
The explanations are two-fold:
On one hand, the ResNet itself is a compact-designed network, there might exist less redundant parameters.
On the other hand, ILSVRC-2012 is a large-scale dataset and contains 1,000 categories, which is much complex than the small-scale CIFAR-10 with only 10 categories.
Second observation comes that ABCPruner obtains higher pruning rates and less accuracy drops as the depth of network increases.
To explain, compared with shallow ResNets, \emph{e.g.}, ResNet-18/34, the deeper ResNets, \emph{e.g.}, ResNet-50/191/152 contain relatively more redundancies, which means more pointless parameters are automatically found and removed by ABCPruner.

%
%

\begin{figure}[!t]
\begin{center}
\includegraphics[height=0.35\linewidth]{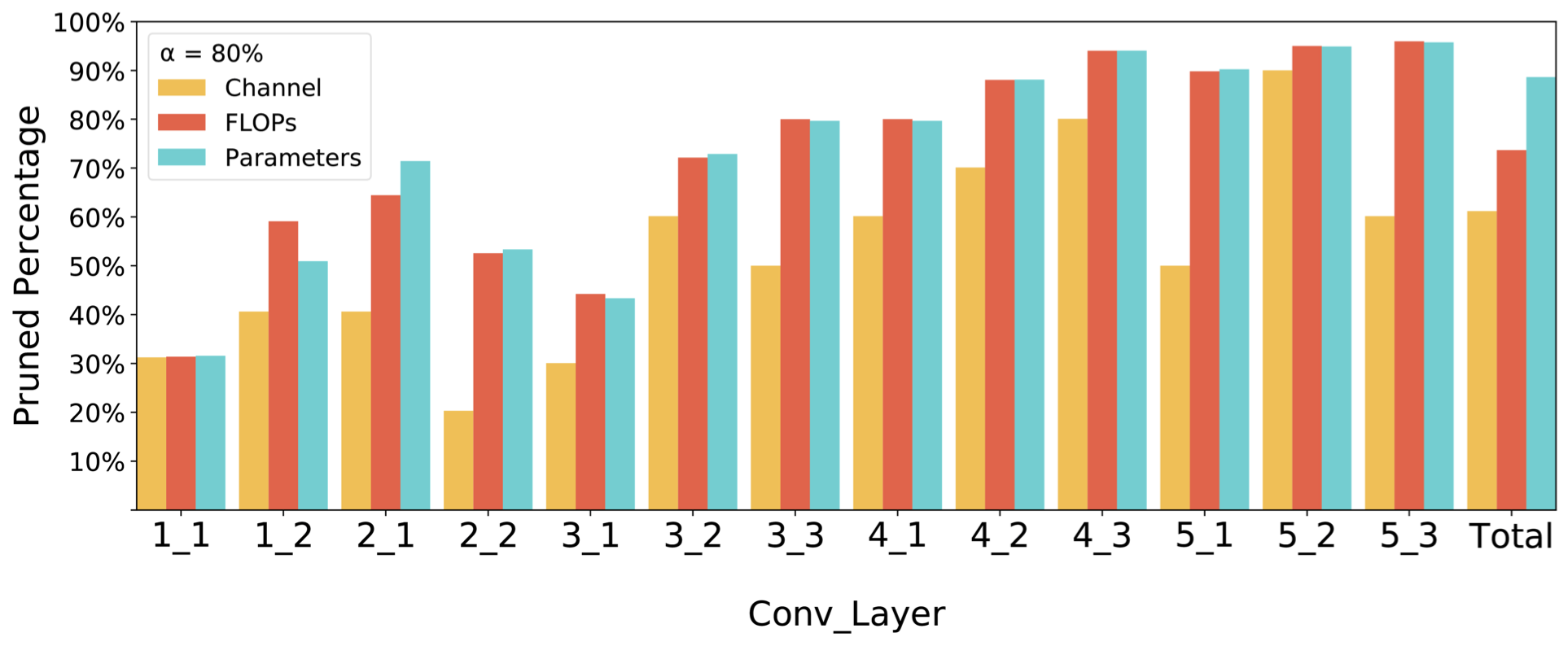}
\end{center}
\caption{\label{ratio}
The pruned percentage of each layer for VGGNet on CIFAR-10 when $\alpha = 80\%$.
}
\end{figure}

\paragraph{In-depth Analysis.}
Combining Tab.\,\ref{cifar} and Tab.\,\ref{imagenet}, we can see that ABCPruner can well compress popular CNNs while keeping a better or at least comparable accuracy performance against the full model.
Its success lies in the automatically optimal pruned structure search.
For deeper analysis, we display the layer-wise pruning results by ABCPruner-80\% for VGGNet-16 in Fig.\,\ref{ratio}.
ABCPruner-80\% denotes that at most 80\% percentage of the channels are preserved in each layer.
As can be seen, over 20\% channels are removed for most layers and the pruning rate differs across different layers.
Hence, the ABCPruner can automatically obtain optimal pruned structure which then feeds back good performance.

\subsection{Comparison with Other Methods}\label{comparison}
In Tab.\,\ref{other}, we compare ABCPruner with traditional methods \cite{luo2017thinet,he2017channel,huang2018data,lin2019towards}, and automatic method \cite{liu2019metapruning}\footnote{We have carefully fixed the codes provided by the authors and re-run the experiments using our baseline of ResNet-50.}.
%

%
\paragraph{Effectiveness.}
The results in Tab.\,\ref{other} show that ABCPruner obtains better FLOPs reduction and accuracy performance.
Compared with importance-based pruning \cite{luo2017thinet,he2017channel} and regularized-based pruning \cite{huang2018data,lin2019towards}, ABCPruner also advances in its end-to-end fine-tuning and automatic structure search.
It effectively verifies the finding in \cite{liu2019rethinking} that optimal pruned structure is more important in channel pruning, rather than selecting ``important" channels.
In comparison with the automatic-based MetaPruning \cite{liu2019metapruning}, the superiority  of ABCPruner can be attributed to the one-stage design where the fitness of pruned structure is directly measured in the target network without any additionally introduced network as in \cite{liu2019metapruning}.
Hence, ABCPruner is more advantageous in finding the optimal pruned structure.
%

\begin{table}[]
\small
\centering
\setlength{\tabcolsep}{0.2em}
\begin{tabular}{|c|c|c|c|c|}
\hline
Model           &FLOPs &Top1-acc         &Baseline-acc  &Epochs\\ \hline

ThiNet-30     & 1.10G  & 68.42\% &76.01\% &244 (196 + 48)\\
\textbf{ABCPruner-30\%} &0.94G &70.29\%  &76.01\% &102 (12+90)\\
SSS-26         &2.33G   &71.82\%  &76.01\% &100\\
GAL-0.5       & 2.33G  & 71.95\%  &76.01\% & 150 (90 + 60)\\
GAL-0.5-joint & 1.84G  &71.80\% &76.01\%  &150 (90 + 60)\\
ThiNet-50     & 1.71G  & 71.01\%  &76.01\%  &244 (196 + 48)\\
\textbf{ABCPruner-50\%} &1.30G &72.58\%  &76.01\%  &102 (12+90)\\
SSS-32         &2.82G   &74.18\%  &76.01\%  &100\\
CP            & 2.73G  & 72.30\% &76.01\%   &206 (196 + 10)\\
\textbf{ABCPruner-100\%}&2.56G &74.84\% &76.01\%  &102 (12+90)\\ \hline

MetaPruning-0.50 &1.03G  &69.92\%  &76.01\%  & 160 (32 + 128)\\
\textbf{ABCPruner-30\%} &0.94G &70.29\% &76.01\%  &102 (12+90)\\
MetaPruning-0.75 & 2.26G  & 72.17\%  &76.01\%  &160 (32 + 128)\\\
\textbf{ABCPruner-50\%} &1.30G &72.58\% &76.01\%  &102 (12+90)\\
MetaPruning-0.85 & 2.92G  & 74.49\%  &76.01\%  &160 (32 + 128)\\\
\textbf{ABCPruner-100\%} &2.56G  &74.84\% &76.01\%  &102 (12+90)\\  \hline
\end{tabular}
\caption{\label{other}The FLOPs reduction, top-1 accuracy and training efficiency of ABCPruner and SOTAs including ThiNet \protect\cite{luo2017thinet}, CP \protect\cite{he2017channel}, SSS \protect\cite{huang2018data}, GAL \protect\cite{lin2019towards}, and MetaPruning \protect\cite{liu2019metapruning} for ResNet-50 \protect\cite{he2016deep} on ILSVRC-2012.}
\end{table}

\paragraph{Efficiency.}
We also display the overall training epochs for all methods in Tab.\,\ref{other}.
As seen, ABCPruner requires less training epochs of 102 compared with others.
To analyze, 12 epochs are used to search for the optimal pruned structure and 90 epochs are adopted for fine-tuning the optimal pruned network.
We note that the importance-based pruning methods \cite{luo2017thinet,he2017channel} are extremely inefficient since they require layer-wise pruning or optimization (196 epochs).
Besides, 48 epochs \cite{luo2017thinet} and 10 epochs \cite{he2017channel} are used to fine-tune the network.
\cite{huang2018data} consume similar training time with ABCPruner, but suffers more accuracy drops and less FLOPs reduction.
\cite{lin2019towards} cost 90 epochs for retraining the network and additional 60 epochs are required for fine-tuning.
Besides, ABCPruner also shows its better efficiency against the automatic-based counterpart \cite{liu2019metapruning} which is a two-stage method where 32 epochs are first adopted to train the large PruningNet and 128 epochs are used for fine-tuning.
Hence, ABCPruner is more efficient compared to the SOTAs.

%
%
%

\subsection{The Influence of $\alpha$}\label{alpha_influence}
\begin{figure}[!t]
\begin{center}
\includegraphics[height=0.4\linewidth]{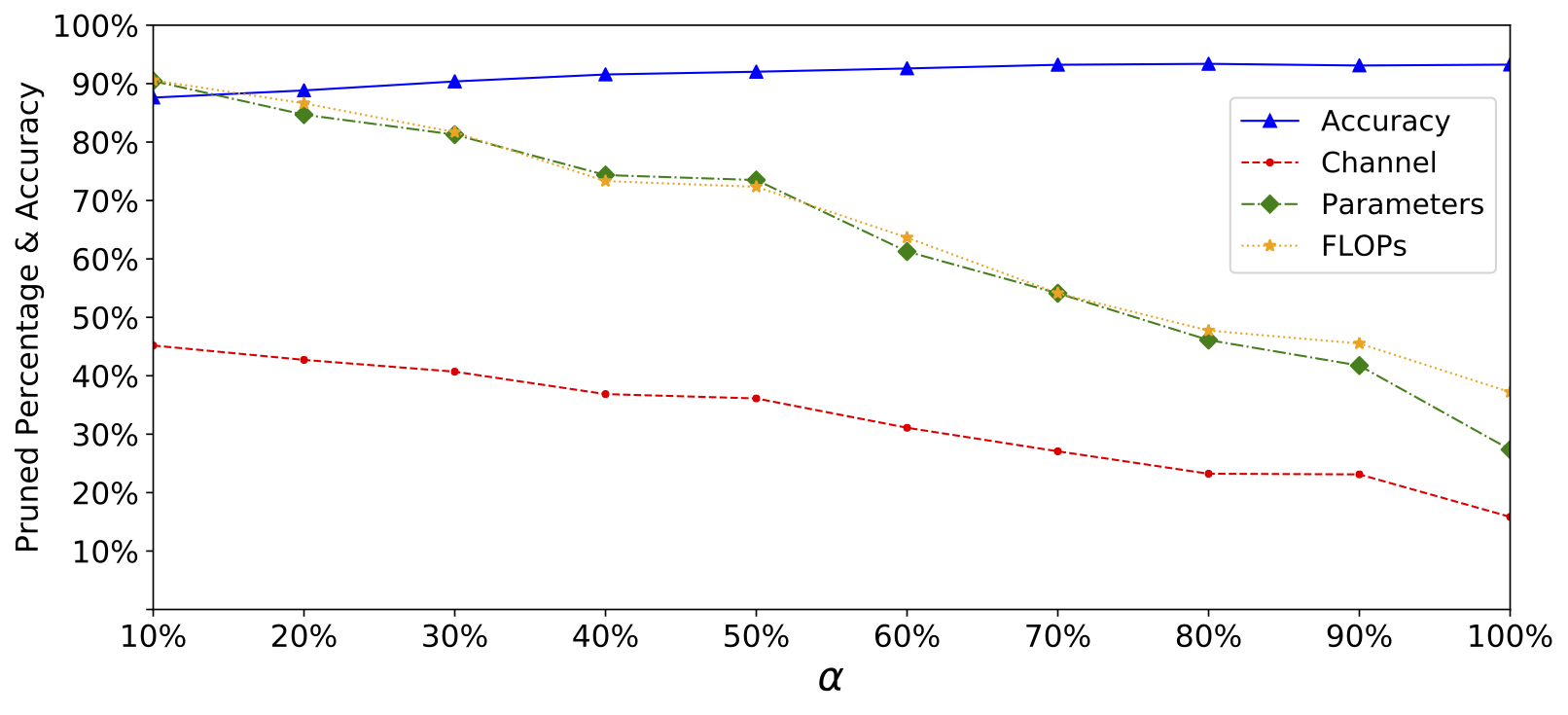}
\end{center}
\caption{\label{alpha}
The influence of $\alpha$ for ResNet-56 on CIFAR-10.
Generally, large $\alpha$ results in better accuracy but less pruning rate.
}
\end{figure}

In this section, we take ResNet-56 on CIFAR-10 as an example and analyze the influence of the introduced constant $\alpha$, which represents the upbound of preserved channel percentage in each layer.
It is intuitive that larger $\alpha$ leads less reductions of channel, parameters and FLOPs, but better accuracy performance.
The experimental results in Fig.\,\ref{alpha} demonstrate this assumption.
To balance the accuracy performance and model complexity reduction, in this paper, we set $\alpha$ = 70\% as shown in Tab.\,\ref{cifar}.

\section{Conclusion}
In this paper, we introduce a novel channel pruning method, termed as ABCPruner.
ABCPruner proposes to find the optimal pruned structure, \emph{i.e.}, channel number in each layer via an automatic manner.
%
We first propose to shrink the combinations of pruned structure, leading to efficient search of optimal pruned structure.
Then, the artificial bee colony is integrated to achieve the optimal pruned structure search in an automatic manner.
Extensive experiments on popular CNNs have demonstrated the efficacy of ABCPruner over traditional channel pruning methods and automatic-based counterpart.

\section*{Acknowledgements}
This work is supported by the Nature Science Foundation of China (No.U1705262, No.61772443, No.61572410, No.61802324 and No.61702136), 
National Key R\&D Program (No.2017YFC0113000, and No.2016YFB1001503), and Nature Science Foundation of Fujian Province, China (No. 2017J01125 and No. 2018J01106).

\bibliographystyle{named}
\bibliography{main}
\end{document}